\documentclass[letterpaper, 10 pt, conference]{ieeeconf}  

\IEEEoverridecommandlockouts                              

\usepackage{graphics}
\usepackage[caption=false]{subfig}
\usepackage{graphicx}
\usepackage{balance}

\usepackage{cite}

\usepackage{amsmath}
\usepackage{amssymb}  

\usepackage{amsthm}  
\usepackage{algorithm}
\usepackage{algpseudocode}
\usepackage{bm}
\algdef{SE}[DOWHILE]{Do}{doWhile}{\algorithmicdo}[1]{\algorithmicwhile\ #1}%

\usepackage{lipsum}
\usepackage{todonotes}
\usepackage{balance}

\usepackage{mathtools}

\DeclarePairedDelimiterX{\infdivx}[2]{(}{)}{%
  #1\;\delimsize\|\;#2%
}

\usepackage{siunitx}

\newcommand{\transpose}{^\top}

\newcommand{\prm}[0]{PRM$^*$}

\newcommand{\fmt}[0]{FMT$^*$}

\newcommand{\trim}[0]{\boldsymbol{\tau}}
\newcommand{\Trim}[0]{\mathcal{T}}

\newcommand{\pos}[0]{\mathbf{p}}
\newcommand{\Pos}[0]{\mathbf{P}}

\newcommand{\argmin}[1]{\underset{#1}{\arg\min}~}
\newtheorem{problem}{Problem}

\newtheorem{remark}{Remark}

\title{\LARGE \bf
Streamline-Based Control of Underwater Gliders in 3D Environments
}

\author{K. Y. Cadmus To$^1$,
        James Ju Heon Lee$^1$,
        Chanyeol~Yoo$^1$,
        Stuart Anstee$^2$
        and Robert~Fitch$^1$
\thanks{This research is supported by an Australian Government Research Training Program (RTP) Scholarship, Australia's Defence Science and Technology Group, and the University of Technology Sydney.}
\thanks{$^{1}$Authors are with the University of Technology Sydney,  Australia
    {\tt\footnotesize \{Cadmus.To,JuHeon.Lee\}@student.uts.edu.au} and {\tt\footnotesize \{Chanyeol.Yoo,Robert.Fitch\}@uts.edu.au}}
\thanks{$^{2}$Author is with the Defence Science and Technology Group, Department of Defence, Australia
    {\tt\footnotesize stuart.anstee@dst.defence.gov.au}}%
}

\begin{document}

\maketitle
\thispagestyle{empty}
\pagestyle{empty}

\begin{abstract}
Autonomous underwater gliders use buoyancy control to achieve forward propulsion via a sawtooth-like, rise-and-fall trajectory.
Because gliders are slow-moving relative to ocean currents, glider control must consider the effect of oceanic flows.
In previous work, we proposed a method to control underwater vehicles in the (horizontal) plane by describing such oceanic flows in terms of streamlines, which are the level sets of stream functions. 
However, the general analytical form of streamlines in 3D is unknown.
In this paper, we show how streamline control can be used in 3D environments by assuming a 2.5D model of ocean currents.
We provide an efficient algorithm that acts as a steering function for a single rise or dive component of the glider's sawtooth trajectory, integrate this algorithm within a sampling-based motion planning framework to support long-distance path planning, and provide several examples in simulation in comparison with a baseline method.
The key to our method's computational efficiency is an elegant dimensionality reduction to a 1D control region.
Streamline-based control can be integrated within various sampling-based frameworks and allows for online planning for gliders in complicated oceanic flows.
\end{abstract}

\section{Introduction}

Underwater gliders are energy-efficient autonomous underwater vehicles~(AUVs) with many important applications in oceanography, industry, and defence.
Examples include monitoring and surveying~\cite{Rudnick2004}, oil and gas exploration~\cite{Russell-Cargill2018}, and underwater surveillance~\cite{Johannsson2010}.
Underwater gliders travel slowly to minimise their energy consumption; their forward velocity can be much lower than the prevailing current, and thus successful navigation depends on finding efficient routes that use oceanic flows to maximum advantage.
A unique characteristic of gliders is that they achieve forward propulsion through a sawtooth-like motion (a sequence of rises and falls) that results from active control of buoyancy~\cite{lee2017energy}.
We recently proposed an efficient control scheme for underwater vehicles in the horizontal plane that is based on the concept of \emph{stream functions} in fluid dynamics, and we are now interested in applying this concept for controlling the 3D motion of underwater gliders.

Our previous work~\cite{cadmus2019streamlines} introduced the idea of superimposing control actions on the level sets of stream functions, known as streamlines.
One set of streamlines arises from the ocean current itself.
Another can be shown to arise from the effect of the AUV's control surfaces in still water.
Superimposing these two sets of streamlines gives the path that an AUV would follow, given the estimated ocean current and control action.
We showed how to efficiently find a control action to produce a streamline that steers the AUV to a desired location.
This control scheme is remarkably simple and supports fast planning over long distances.

Applying streamlines in 3D, however, is not straightforward since the analytical form of general stream functions in 3D is not known.
Further, although ocean currents change continuously with depth, good 3D estimates are not readily available.
Continuous current models can be constructed from non-uniformly collected data, but widely available current predictions are generally provided at discrete depth values.
Some form of approximation is thus necessary to model ocean currents within a 3D ocean volume.

In this paper, we consider a 2.5D ocean current model and focus on developing a streamline-based \emph{steering function} for underwater gliders in this setting.
This problem is important for the development of long-range motion planning algorithms because it allows computation of edge connectivity in sampling-based planners, a common class of motion planning algorithms in robotics~\cite{LaValle2006}.
We consider steering functions that encompass a single rise or fall component of the sawtooth trajectory.
The idea is that a sequence of such rises and falls will be concatenated to generate a long-distance motion plan.
The glider will generally not follow a straight-line path due to the interaction of its control surfaces with the current.

The 2.5D model assumes that the vertical component of current is negligible.
We average the horizontal flow locally, between successive control adjustment positions.
This assumption is reasonable for relatively small changes in depth, since ocean current velocity is dominated by its horizontal components by several orders of magnitude~\cite{gliderPP_3D_diffEvol_2014,stewart2008introduction}.
Change in depth is typically limited to a few hundred metres, either by physical constraints of the glider itself or by applications that require surveying within a given depth range.

We present a fast edge evaluation algorithm for the steering functions using streamlines and demonstrate its use using a well-known sampling-based algorithm \emph{probabilistic roadmap$^*$} (\prm)~\cite{star_Karaman2010}. 
Instead of exhaustively sampling over a set of controls that satisfy glider dynamics and then forward integrating, we approximate the streamlines based on the 2.5D current model.
The streamline approach allows us to reduce the dimensionality of the overall control space from a (2D) surface to a parameterised (1D) line.
This dimensionality reduction allows for dense sampling, improving the likelihood of finding a solution.

We integrate our algorithm with \prm{} and demonstrate two simulated examples.
We show that the overall framework is able to find a solution with fewer control samples and higher path quality than a baseline method.
Note that most sampling-based planning methods would benefit from the proposed method since the major computational bottleneck is in finding the reachability between two points.
The significance of this result is that glider path planning can be performed efficiently in an online manner, enabling long-duration autonomous deployments without the need for frequent manual intervention.

\section{Related work}
From a control perspective, the problem of navigating in flow field has been well-studied~\cite{oneHalfTurnRate_Techy_2011,efficientGuidance_1D_Rhoads_2013,Touring_twoVortx_Lagor_2015,navi_reducedOrder_Lorenz_2015}.
The most relevant to our problem is the minimum time feedback control~\cite{minTimeFeedbackCtrl_HJB_Rhoads_2010} that solved for optimal feedback control law from the dynamic Hamilton Jacobi Bellman~(HJB) equation.
However such methods only consider direct velocity control, while control over all states of a common underwater glider such as the SLOCUM~\cite{webb_slocumDynamics_2001,Jones_SlocumInfo_2014} would be computationally prohibitive.

Our approach for optimal path planning in a flow field is similar to using level-set methods to find time-optimal~\cite{Lolla2014a} and energy-optimal paths~\cite{subramani2016energy}, where the vehicle's reachable sets are explicitly solved as a level set.
The level sets are solved using partial differential equations, making it computationally expensive for robotics applications.

Other planning methods include a variety of graph-based methods where the workspace is discretised uniformly~\cite{Kularatne2012}, adaptively~\cite{Kularatne2018c}, or biased to the flow field~\cite{gliderPP_CTS_A*_2010,Huynh_NRMPC_2015}.
These algorithms are resolution-complete, where there is a trade-off between the performance and solution quality. 
Sampling-based methods such as RRT~\cite{LaValle2001} have also been considered~\cite{dushyantGlider_2009, VF-RRT_2014}, including VF-RRT~\cite{VF-RRT_2014} that biases its samples with the flow field.
In our previous work, we presented a \fmt{}-based planning algorithm in a 3D flow field in various ways~\cite{lee2017energy,brian2018plume}.
In~\cite{lee2017energy}, we produced asymptotically optimal minimum energy path with trim-based controls.
The major bottleneck we encountered was from the edge cost evaluation that needed to iterate through all valid controls.

Aside from our previous work, studies on underwater glider path planning in 3D flow fields have been surprisingly rare.
A large portion of existing work considers a 2D oceanic flow either using surface currents~\cite{gliderPP_LCS_2005,dushyantGlider_2009} or implicitly using a depth-average current~\cite{gliderPP_CTS_A*_2010,singleMulti_gliderPlanning_2011,geneticAlg_glider_2017}.
The few that do plan in 3D workspace either model the glider with simple kinematics and a directly controllable turning rate~\cite{gliderPP_3D_diffEvol_2014}, or do not account for flow fields~\cite{Liu_3Dglider_controlConstraint_2017,glider_3DPlanning_2017}.

In this paper, we propose a method to reduce a 2D control search space to 1D in a 2.5D flow field.
This work is an extension of our previous work~\cite{cadmus2019streamlines}, which only dealt with 2D flow fields.

\section{Background}
\subsection{Trim-based control of underwater gliders}\label{subsec:bkgd_gliderCtrl}
In this section we describe how to control an underwater glider using a sequence of fixed controls, where the underlying dynamic model is reduced to a kinematic model.
The dynamics of a glider~$G$ is modelled as
\begin{equation} \label{eqn:dynamical}
    \dot{\mathbf{x}} = f(\mathbf{x}, \mathbf{u}) + \mathbf{v}_c(\mathbf{x})
    ,
\end{equation}
where~$\mathbf{x}$ is the 12D glider state in the~$\mathbb{R}^3$ workspace~\cite{leonard_gliderDynamics_2001},
$\mathbf{u}$ is the control input, 
and ${\mathbf{v}_c(\mathbf{x}) = [u_c,v_c,w_c,0,\ldots,0]\transpose}$ is the time-invariant ocean flow vector in $xyz$ space at the glider's position.
We express the flow velocity $\mathbf{v}_c(\mathbf{x})$ in state space with the non-velocity dimensions set to zero.
We consider the vertical flow $w_c$ to be zero, in line with common practice~\cite{gliderPP_3D_diffEvol_2014,dushyantGlider_2009}.

The glider gains forward velocity by shifting its centre of mass and pumping water in or out from a ballast tank to change its buoyancy.
The control input~${\mathbf{u}(t) = [\mathbf{u}_{\mathbf{r}_p}(t), \mathbf{u}_{mb}(t)]\transpose}$ consists of the force applied to adjust the centre of mass~$\mathbf{u}_{\mathbf{r}_p}(t)$ and the rate of water inflow to the ballast tank~$\mathbf{u}_{mb}(t)$. 

Since the glider is designed to be energy conservative and much of the energy expenditure comes from changing control inputs,
controlling the glider continuously over time is not desirable.
In this paper, we exploit \emph{trim state} manoeuvres to reduce the dynamic problem to a kinematic problem.
A trim state is the state of dynamic equilibrium that is maintained under no disturbance or control variation~\cite{leonard_gliderDynamics_2001,lee2017energy}.

A glider at ${\mathbf{p}_k = [x_k,y_k,z_k]\transpose}$ has trim state $\trim_k$ as:
\begin{equation} \begin{split} \label{eqn:pos_state}
    \trim_k &= 
    \begin{bmatrix}
    V_{G,k}	&	\gamma_k	&	\delta_k	&	m_{b,k}
    \end{bmatrix}\transpose
    ,
\end{split} \end{equation}
where~${\gamma_k \in \Gamma = [\gamma_{min},\gamma_{max}] \cup [-\gamma_{max}, -\gamma_{min}]}$ is the glide angle bounded by both glider constraints and safety factors~\cite{lee2017energy},
$\delta_k$ is heading angle,
$m_{b,k}$ is the ballast mass that we assume is either empty or full ($m_{b,k} \in \{0, m_{b\max}\}$),
and $V_{G,k}$ is the glider speed, which is a nonlinear function of $\gamma_k$ and $m_{b,k}$:
\begin{equation} \label{eqn:gliderVel}
    V_{G,k}(\gamma_k, m_{b,k}) = \sqrt{ \frac{ (m_{b,k} - (m - m_h - \bar{m}))\cdot g }{ -D(\gamma_k) \sin{\gamma_k} + L(\gamma_k) \cos{\gamma_k}}}
    ,
\end{equation}
where $D(\gamma)$ is the drag force, $L(\gamma)$ is the lift force,
and $g$ is the acceleration due to gravity. The rest of the constants are explained in~\cite{lee2017energy}.
Note that due to physical constraints, both glider speed~$V_{G,k}$ and ballast mass~$m_{b,k}$ can be strictly formulated as a function of~$\gamma$.
Therefore, a trim state~$\trim$ is a function of glide and heading angles.

Given the glide angle~$\gamma_k$ and heading angle~$\delta_k$, the glider velocity vector or \emph{control vector} in $xyz$ space is
\begin{equation} \begin{split} \label{eqn:gliderDynamic_model}
    \begin{bmatrix}
        u_{G,k}	\\
        v_{G,k}	\\
        w_{G,k}
    \end{bmatrix}(\gamma_k,\delta_k) = 
    \begin{bmatrix}
        V_G(\gamma_k) \cos \gamma_k \cos \delta_k	\\
        V_G(\gamma_k) \cos \gamma_k \sin \delta_k	\\
        V_G(\gamma_k) \sin \gamma_k
    \end{bmatrix}
    .
\end{split} \end{equation}
The set of such control vectors are referred to as the \emph{control surface}.
\begin{remark} [Glider control surface] \label{remark:control_surface}
    Given the set of all valid glide angles $\gamma_k$ and heading angles $\delta_k$ that satisfy the constraints on glider $G$, the glider control surface is the set of valid controls (i.e., glider velocities) allowed by the dynamic in~\eqref{eqn:gliderDynamic_model}.
\end{remark}

A sequence of trim states or \emph{trim sequence} is:
\begin{equation}
   \Trim = \trim_0\trim_1\cdots\trim_{K-1},
\end{equation}
where $K$ is the number of trim states.
We denote the transition cost from position~$\mathbf{p}_k$ to~$\mathbf{p}_{k+1}$ as $c(\mathbf{p}_k, \mathbf{p}_{k+1})$.
We implicitly denote~$\trim_k$ as the cost-minimal trim state from~$\mathbf{p}_k$ to~$\mathbf{p}_{k+1}$ and denote the overall cost for trim sequence~$\Trim$ from~$\mathbf{p}$ as $c_{\Trim}(\mathbf{p})$.

\begin{figure}
	\centering
	\subfloat[Example 2D streamlines]{
        \includegraphics[width=0.41\columnwidth]{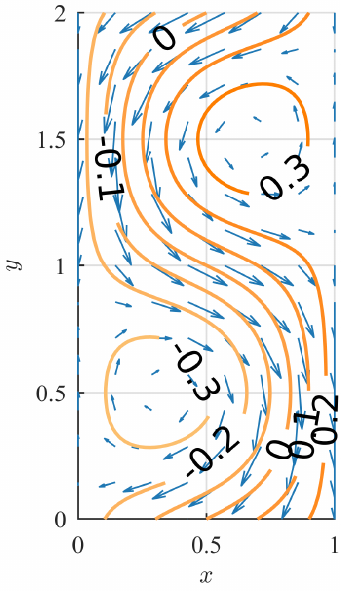}\label{fig:streamlines}
    }
	\subfloat[Trajectories from fixed controls]{
        \includegraphics[width=0.51\columnwidth]{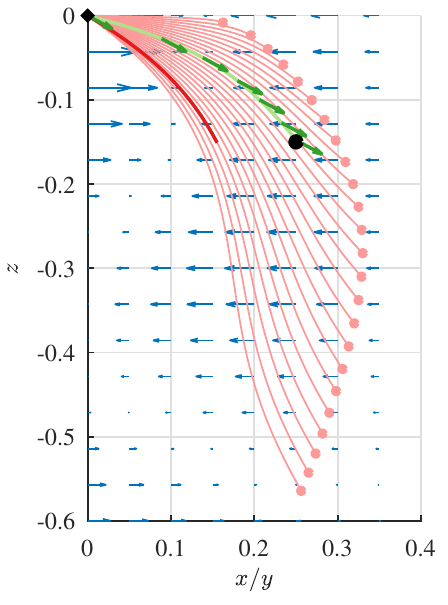}\label{fig:gliderTrajComparison}
    }
	\caption{
        \protect\subref{fig:streamlines} Streamlines (orange lines) can be represented as disjoint contour lines of the stream function of a 2D incompressible flow field (blue arrows).
        Each streamline is labelled with their stream values relative to the reference point $(0.5,1.0)$.
        \protect\subref{fig:gliderTrajComparison} Forward integration with \eqref{eqn:dynamical} is necessary to determine reachability from the glider (diamond) to the target (circle).
        The trajectories (pink lines) of different velocities from the control space~\eqref{eqn:gliderDynamic_model} are used to determine the best control to find a trajectory that can reach the target (green line).
        The naive assumption that currents are uniform leads to a very different trajectory (darker red line).
    }
\end{figure}

\subsection{Streamline-based control search in 2D flow fields} \label{subsec:streamline-basedCtrlSearch}

In our previous work~\cite{cadmus2019streamlines}, we used \emph{stream functions} to find a fixed control~$[u_{G,k}, v_{G,k}]\transpose$ in a 2D horizontal plane to traverse from position~${\mathbf{p}_k = [x_k,y_k]\transpose}$ to~$\mathbf{p}_{k+1}$ in the presence of flow field~${\mathbf{v}_c(\mathbf{p}) = [u_c, v_c]\transpose(\mathbf{p})}$.
This method reduced the sampled control space, leading to faster computation for the same path quality.

Given that oceanic flow is incompressible~(i.e $\nabla\cdot\mathbf{v}_c = 0$),
a \emph{stream function} $\psi: \mathbb{R}^2\times\mathbb{R}^2 \rightarrow\mathbb{R}$ can be defined as~\cite{Batchelor1967}
\begin{equation} \label{eqn:streamValue}
    \psi_c(\mathbf{p}_k,\mathbf{p}_{k+1}) = \int^{\mathbf{p}_{k+1}}_{\mathbf{p}_k} (u_c(\mathbf{p})dy - v_c(\mathbf{p})dx)
    .
\end{equation}
Stream functions quantify the path-independent net flow flux from one point to another.
We refer to this flux as the \emph{stream value} between two points.

In a 2D environment, a set of continuously joined points with the same stream value relative to an arbitrary point is referred as a \emph{streamline}.
Fig.~\ref{fig:streamlines} illustrates this for different stream values.
Intuitively, a stationary vehicle would drift along a streamline in the direction of the flow.

Since two stream functions can be added together, we form superimposed stream functions by combining the stream function of the flow field and that due to glider control.
Intuitively, a glider with a given control would follow one of the superimposed streamlines.
We noted in our previous work~\cite{cadmus2019streamlines} that a vehicle can only reach points with stream values of zero relative to the starting point.
This constrains the possible controls that can be chosen.
We refer to this constraint as the \emph{streamline constraint}.
\begin{remark} [Streamline constraint] \label{remark:streamline_constraint}
    Given the stream function of a 2D incompressible flow field $\psi_c(\mathbf{p},\mathbf{p}^\prime)$, a vehicle with constant velocity~$[u_{G,k}, v_{G,k}]\transpose$ at starting position $\mathbf{p}_k$ and a goal position $\mathbf{p}_{k+1}$, a vehicle satisfies a streamline constraint only if
    \begin{equation} \label{eqn:gliderVelocityline}
        \psi_c(\mathbf{p}_{k},\mathbf{p}_{k+1}) + \psi_G(\mathbf{p}_{k},\mathbf{p}_{k+1}) = 0
        ,
    \end{equation}
    where 
    \begin{equation}
        \psi_G(\mathbf{p}_{k},\mathbf{p}_{k+1}) = (y_{k+1}-y_{k})u_{G,k} - (x_{k+1}-x_{k})v_{G,k}
    .
    \end{equation}
\end{remark}

In 2D flow fields, the controls that satisfy the stream constraint lie along a straight line over~$u_G$ and~$v_G$, which we call the \emph{control line}.
\begin{remark} [Control line] \label{remark:control_line}
    Given the stream function of a 2D incompressible flow field $\psi_c(\mathbf{p},\mathbf{p}^\prime)$, a vehicle at starting position $\mathbf{p}_k$ and a goal position $\mathbf{p}_{k+1}$, the set of fixed control vectors~$\ell_{k, k+1}$ that takes the vehicle to the goal forms a straight line in control space, such that
    \begin{equation}
        \ell_{k,k+1} = \{(u_{G,k},v_{G,k}) \mid  \psi_c(\mathbf{p}_{k},\mathbf{p}_{k+1}) + \psi_G(\mathbf{p}_{k},\mathbf{p}_{k+1}) = 0\}
        .
    \end{equation}
\end{remark}
\noindent Essentially, the control line is a constraint on the fixed controls the vehicle can take to reach its destination.
It reduces the search space for a control to reach the goal from 2D to 1D, allowing fewer control samples to be used to find a suitable control and reducing overall computation time.

\section{Problem statement}
We now consider a path planning problem in a 3D environment in which an autonomous underwater glider~$G$ moves under the influence of an incompressible flow field~$\mathbf{v}_c$, and the glider control is trim-based as shown in Sec.~\ref{subsec:bkgd_gliderCtrl}. In contrast to our previous work, in this case the glider is able to reverse its direction of motion (upward or downward) at arbitrary depths and it is also able to adjust its glide angle whenever it changes trim.

\begin{problem} [Trim-based path planning for underwater glider in flow field]
    Given glider $G$, initial position $\pos_{init}$, goal position $\mathbf{p}_{goal}$, and an incompressible flow field $\mathbf{V}_c$, find the optimal sequence of position vectors~$\mathbf{P}^* = \pos_0 \mathbf{p}_1 \cdots$ that minimises the overall cost to traverse from $\pos_{init}$ to $\pos_{goal}$ such that
    \begin{equation}
        \Pos^* = \argmin{\Pos} \sum^{K-2}_{k=0} c(\mathbf{p}_k, \mathbf{p}_{k+1})
        ,
    \end{equation}
    where~$\mathbf{p}_0 = \pos_{init}$ and $\pos_{K-1} = \pos_{goal}$.
\end{problem}
\noindent The sequence of position vectors~$\mathbf{P}^*$ is found by computing the trim states between consecutive position vectors within the sequence.
The sequence of such trim states is denoted as~$\Trim^* = \trim^*_0 \cdots \trim^*_{K-2}$.

Finding a trim state connecting two consecutive position vectors in a flow field is hard~\cite{leonard_gliderDynamics_2001,lee2017energy}.
This boundary value problem is known as \emph{Zermelo's problem}~\cite{Zermelo_RefBook1931}, and it has no known analytical solution.
Therefore, in the worst case, trim states can only be found by exhaustively sampling the set of control vectors that satisfies the glider dynamics which we describe in Remark~\ref{remark:control_surface}.
Each control must then be forward integrated from~$\pos_k$ over time horizon~$H$, allowing the choice of a control that reaches~$\pos_{k+1}$ within tolerance.
This problem is even harder with non-linear glider dynamics in 3D environments such as the one illustrated in Fig.~\ref{fig:gliderTrajComparison}.


\begin{figure}[t!]
	\centering
	\subfloat[Isometric view]{
	    \includegraphics[width=0.9\linewidth]{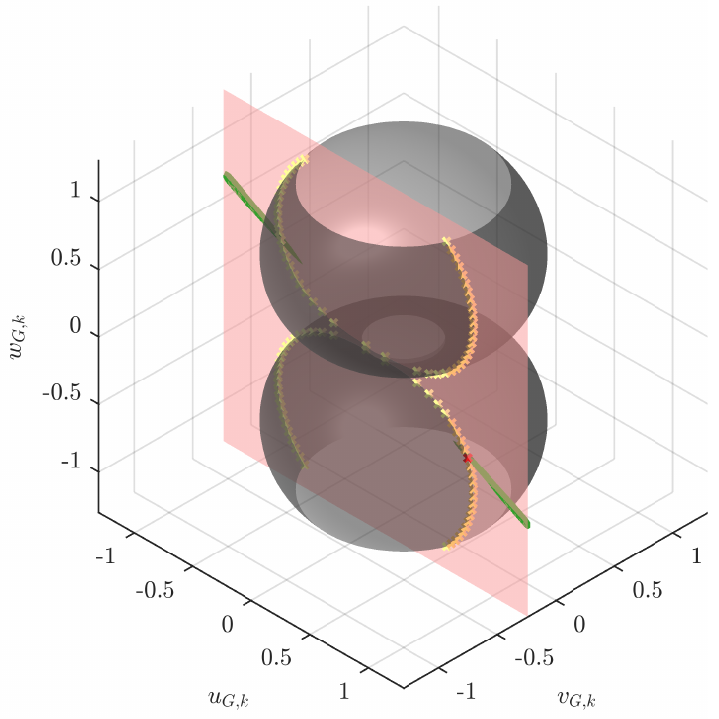}
	}\\
	\subfloat[Top view]{
	    \includegraphics[width=0.9\linewidth]{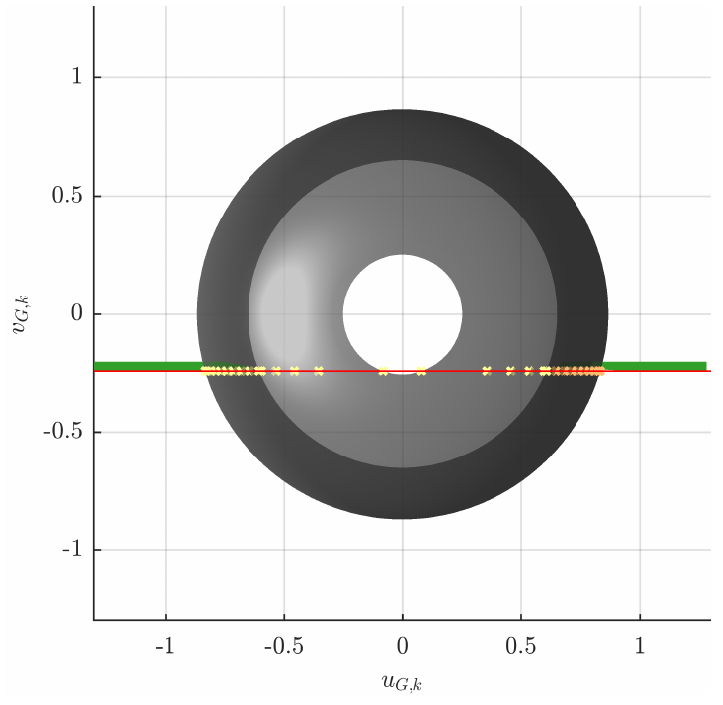}
	}
	\caption{
	    Control surface derived from glider dynamics (grey, Remark~\ref{remark:control_surface}) and control plane derived from the streamline constraint (pink, Remark~\ref{remark:control_plane}) are shown over 3D glider control space.
	    Our method finds a set of control candidates along the parameterised control line (Remark~\ref{remark:param_control_line}) at the intersection of the surface and the plane (yellow crosses).
	    The control candidates using the numerical baseline methods (green volume) lie close to our control plane.
	    The control solution is shown as a red cross.
	}
	\vspace{-2ex}
	\label{fig:gliderVeloSpace}
\end{figure}

\section{Streamline-based control search for a 2.5D flow field}
We present a streamline-based control approach in a 3D environment to find a glider trim state~$\trim_{k+1}$ from position vector~$\pos_k$ to~$\pos_{k+1}$ in the presence of ocean currents~$\mathbf{v}_c$. 
In our previous work, we did not consider the glide angle; instead, the speed of the glider through water was a control parameter. In this case, the planner may choose the range of depths in which the glider moves by inverting the glide sense (up or down) whenever it adjusts the trim.
Rather than sampling controls on the 2D control surface (noted in Remark~\ref{remark:control_surface}),
we present a method to find a parameterised control line along the surface using stream functions.
The parameterised line represents the set of control candidates that satisfies both glider dynamics and the streamline constraint (i.e., stream value between the two position vectors is zero).
This method significantly reduces the computational complexity in finding a solution.

\subsection{2D control plane approximation in a 3D flow field}\label{subsec:bkgd_2Dstreamline}
Suppose we have two positions~$\pos_k \in \mathbb{R}^3$ and~$\pos_{k+1} \in \mathbb{R}^3$, between which ocean currents~$\mathbf{v}_c$ vary spatially.
Without loss of generality, we assume the positions have different depths (i.e., $z_k \not= z_{k+1}$) since glider dynamics normally prohibit a manoeuvre at the same depth. 

A trim state~$\trim_k$ can connect the positions if the corresponding control vector~${[u_{G}, v_{G}, w_{G}]\transpose}$ satisfies the glider dynamics in the presence of ocean currents and reaches the goal position within some tolerance.
This implies that any control candidate should be on the control surface (noted in Remark~\ref{remark:control_surface}) which is illustrated in Fig.~\ref{fig:gliderVeloSpace}.
Note that a control on the surface is computed with~\eqref{eqn:gliderDynamic_model}, which yields the nominal glider speed~$V_G$ for a given pair of glide angle~$\gamma$ and heading~$\theta$.

In order to reduce the complexity in the sampling of controls, we wish to extend the 2D streamline-based approach in Sec.~\ref{subsec:streamline-basedCtrlSearch} to find a 3D control vector, such that the stream value of $\pos_{k+1}$ is zero relative to $\pos_k$.
Intuitively, we need to find a set of control candidates in 3D control space as an extension of control line shown in Remark~\ref{remark:control_line}~\cite{cadmus2019streamlines}.
However, an analytical form yielding ~$\psi_c$ and~$\psi_G$ for such a 3D variant is not available.

In this section, we approximate the set of controls given two positions, $\pos_k$ and~$\pos_{k+1}$, and an intervening flow field~$\mathbf{v}_c$.
As we previously assumed, since ocean currents move primarily horizontally, the $z$-component of ocean currents is zero (i.e., $w_c = 0$ for all~$\mathbf{v}_c$)~\cite{gliderPP_3D_diffEvol_2014,stewart2008introduction,dushyantGlider_2009}.

From this assumption, we have two 2D flow fields for~$\pos_k$ and~$\pos_{k+1}$ over planes on different depths, $z_k$ and~$z_{k+1}$, respectively.
The 2D flow field between the depths is approximated by assuming that the flow field varies linearly with depth.
The flow field is then the average of the flow fields for~$\pos_k$ and~$\pos_{k+1}$ across depth:
\begin{equation}
    \hat{\mathbf{v}}_c(\hat{\pos} \mid \pos_{k}, \pos_{k+1}) = \frac{\mathbf{v}_c([\hat{x}, \hat{y}, z_k]\transpose) + \mathbf{v}_c([\hat{x}, \hat{y}, z_{k+1}]\transpose)}{2}
    ,
\end{equation}
where~$\hat{\pos} = [\hat{x}, \hat{y}, \hat{z}]\transpose$.
This approximation can be made since $w_c=0$ and that glider travels monotonically through $z$.
In practice, the flow field does not vary linearly across depth. However, as we increase the number of samples in~\prm, the connection radius reduces and the assumption becomes valid. It is important to note that the solution approaches the optimal as we increase the number of samples.

Once we have the approximate 2D flow field~$\hat{\mathbf{v}}_c$ between the two depths,
we compute the control line~$\ell_{\pos_k \pos_{k+1}}$ (from Remark~\ref{remark:control_line}) with the corresponding stream function~$\hat{\psi}_c$. 
Intuitively, this control line prescribes the set of control candidates over horizontal plane~$[u_G, v_G]\transpose$ that allows two positions, $\pos_k$ and~$\pos_{k+1}$, to have approximately the same stream value.
Since the stream function is valid for all depths between $z_k$ and $z_{k+1}$, the stream value given two 2D positions is also the same across that depth.
Thus the control line can be projected along the $z$-axis of the control space (i.e., $w_G$).
This projection induces what we refer to as a \emph{control plane}, which is shown in Fig.~\ref{fig:gliderVeloSpace} as a flat plane in pink that intersects with the control surface.
\begin{remark} [Control plane] \label{remark:control_plane}
    Given two 3D position vectors~$\pos_k$ and~$\pos_{k+1}$, a control plane is a set of control vectors that ensures that the stream value of $\pos_{k+1}$ relative to $\pos_{k}$ is zero with the estimated stream function $\hat{\psi}_c$.
    The velocities on the control plane satisfy the constraint:
    \begin{equation} \begin{split} \label{eqn:control_plane}
        A \cdot u_{G,k} + B \cdot v_{G,k} + C = 0
        ,
    \end{split} \end{equation}
    where ${A = y_{k+1} - y_k}$, ${B = -(x_{k+1} - x_k)}$, and ${C = \hat{\psi}_{{c}}(\mathbf{p}_{k},\mathbf{p}_{k+1})}$.
\end{remark}

\subsection{Control parameterisation}
The set of control candidates~$[u_{G,k}, v_{G,k}, w_{G,k}]\transpose$ that satisfies both glider dynamics and streamline constraints is found along the intersection of the control surface and control plane.
Some examples of these control candidates are marked as crosses in Fig.~\ref{fig:gliderVeloSpace}.
The set of controls along the intersection is called the \emph{parameterised control line}.
\begin{remark} [Parameterised control line] \label{remark:param_control_line}
    The set of controls can be analytically represented as a parametric function of glide angle~$\gamma_k$, such that
    \begin{equation} \begin{split} \label{eqn:parametrised}
        \begin{bmatrix}
            u_{G,k}	\\ v_{G,k} \\ w_{G,k}
        \end{bmatrix}(\gamma_k) = 
        \begin{bmatrix}
            \frac{-A\cdot C \pm B\sqrt{(A^2 + B^2)(V_{G}(\gamma_k)\cos \gamma_k)^2-C^2}}{(A^2 + B^2)}	\\
            \frac{-B\cdot C \pm A\sqrt{(A^2 + B^2)(V_{G}(\gamma_k)\cos \gamma_k)^2-C^2}}{(A^2 + B^2)}	\\
            V_{G}(\gamma_k) \sin \gamma_k
        \end{bmatrix}
        ,
    \end{split} \end{equation}
    where the corresponding heading angle is~$\delta_k = \arctan \frac{v_{G,k}}{u_{G,k}}$.
\end{remark}
\noindent  As a result, the set of control candidates is found along a 1D curve, as opposed to a 2D surface. 
In Fig.~\ref{fig:gliderVeloSpace}, example control candidates are placed along the intersection (yellow asterisks).
The true set of control candidates, computed numerically without the 2.5D approximation, is represented by green asterisks, where the control solution is marked by a red asterisk.
The result illustrates that our approximate analytical solution is very close to the exact numerical solution.

\subsection{Lowest speed condition}
The set of control candidates is limited to solutions of~\eqref{eqn:parametrised} if they exist.
Conversely, if there exists no intersection between the control surface and the control plane, no solution is possible and computation time can be further reduced, since control sampling and forward integration are not required. 
We define this condition as the \emph{lowest speed condition}.
\begin{remark} [lowest speed condition] \label{remark:low_speed_cond}
    Given two positions, $\pos_k$ and~$\pos_{k+1}$, and flow field~$\mathbf{v}_c$, there exists no solution if the lowest speed on the control plane~$V_{\min}$ (i.e., lowest plane speed) is greater than the maximum horizontal glide speed, such that
    \begin{equation}
        V_{\min}
                > 
        \max_{\gamma \in \Gamma} V_{G}(\gamma) \cos \gamma
        ,
    \end{equation}
    where~$V_{\min} = \min_{u_G, v_G} \sqrt{u_G^2 + v_G^2}$, and $u_G$ and~$v_G$ are along control plane in~\eqref{remark:control_plane}.
\end{remark}
\noindent Intuitively, no solution is possible if the minimum speed allowed by the control plane exceeds the maximum horizontal speed allowed by the glider dynamics.
Note that the maximum horizontal speed over glide angle is constant.

\section{Analysis}\label{sec:analysis}

\begin{figure}
    \centering
    \subfloat[Number of edge connections in \prm{} graph with respect to the number of control candidates]{
        \includegraphics[width=\linewidth]{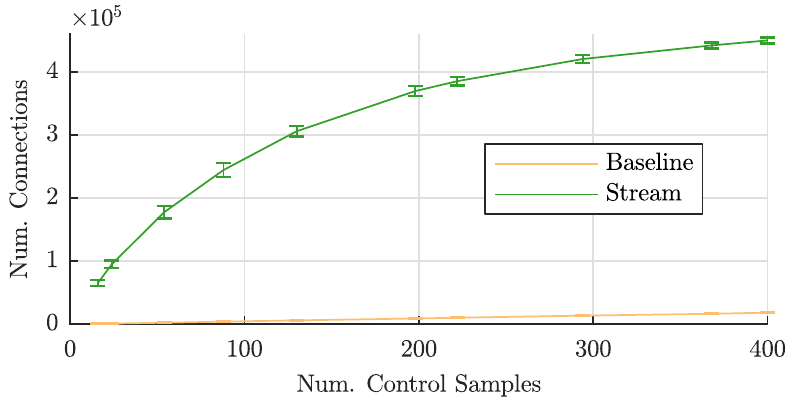}\label{fig:analysis_connectivity}
    }
    \\
    \subfloat[Path quality with respect to the number of control candidates]{
        \includegraphics[width=\linewidth]{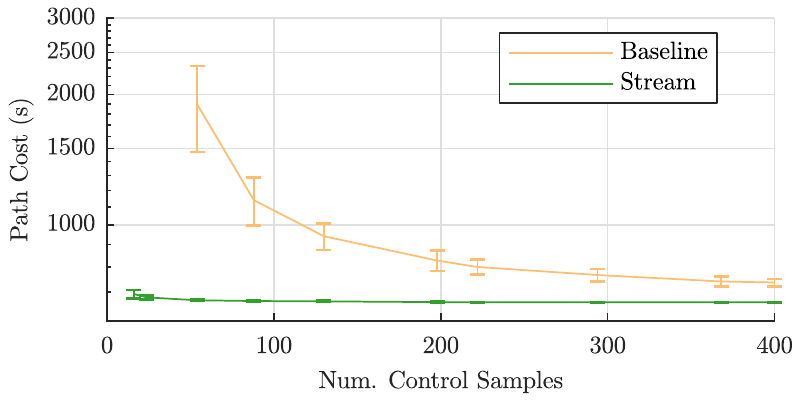}\label{fig:analysis_density}
    }
    \\
    \subfloat[Path quality with respect to \prm{} computation time]{
        \includegraphics[width=\linewidth]{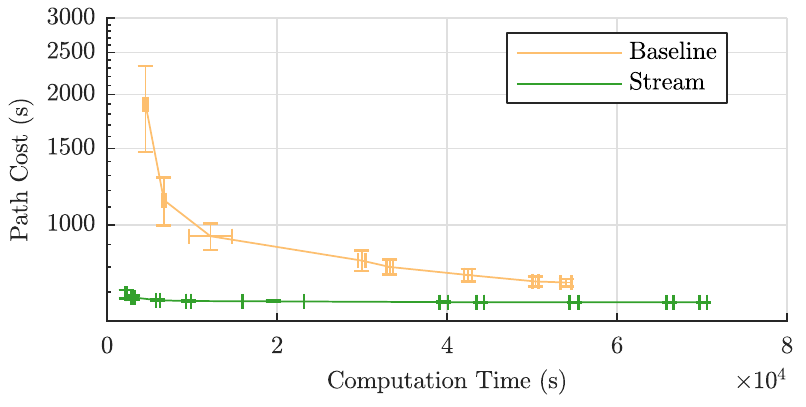}\label{fig:analysis_convergance}
    }
    \caption{
        Performance comparison of the streamline-based and baseline \prm{} (green and orange, respectively) in a 3D flow field with 1000 equispaced position samples in 3D and set of randomly sampled velocities (16 to 400).
        The $99.7\%$ confidence interval over 32 runs around the mean value is also shown.}
    \label{fig:analysis}
\end{figure}



The likelihood of finding a viable control connecting sample positions increases with the number of control samples.
This likelihood can be quantitatively represented using \emph{sample density}~$\rho$ which we define as the average distance between adjacent control samples, such that~$\rho = c / A$, where~$c$ is the number of control samples and $A$ is the size of the control search space.

For the baseline method, controls are sampled from a surface defined by the glider dynamics in~\eqref{eqn:gliderDynamic_model}.
The upper bound for the area of the surface can be found by considering sphere segments with radius~$V_{G\max} = V_G(\gamma_{\max})$ for a valid set of glide angles~$\gamma \in \Gamma$, where~$\gamma_{\max} = \arg\max_{\gamma} V_G(\gamma)$ (i.e., maximum glider speed).
In a big-O notation, the density of the control surface is~$\rho_{s} = \mathcal{O}(\frac{c}{V^2})$ and that of the parameterised control line is~$\rho_{\ell} = \mathcal{O}(\frac{c}{\sqrt{V^2 - V_{\min}^2}})$, where~$V_{\min}$ is the lowest plane speed (as defined in Remark~\ref{remark:low_speed_cond}).
Since the lowest plane speed is determined by a pair of position vectors and~$V_{\min} \in (0, V_{G\max})$, our method yields a much denser set of samples, which finds more edge connections given the same number of control samples. This result is empirically proven in Fig.~\ref{fig:analysis_connectivity}, where the number of connections increases much faster using our method compared to the baseline.

With significantly more edge connections using our method, the path quality converges to the true optimum much quicker than using the baseline, as illustrated in Fig.~\ref{fig:analysis_density}, where our method provides a near-optimal solution with  surprisingly few control samples. As a consequence, the computation time to achieve the same path quality is much less than the time required by the baseline method, as shown in Fig.~\ref{fig:analysis_convergance}. It is also important to note that the baseline method was not able to find any solution with less than 54 control samples while our method found a solution with 16 control samples.

This is an important property to exploit in practical applications. Our method finds significantly better solutions with only a few control samples, thus the overall computation time to find a path can easily be made lower than the path execution time. This implies that our method can be used to re-plan in a \emph{plan-as-you-go} manner.

\begin{figure*}
    \centering
    \begin{minipage}{0.5\textwidth}
        \centering
        \subfloat[Isometric view]{
            \includegraphics[width=0.9\linewidth]{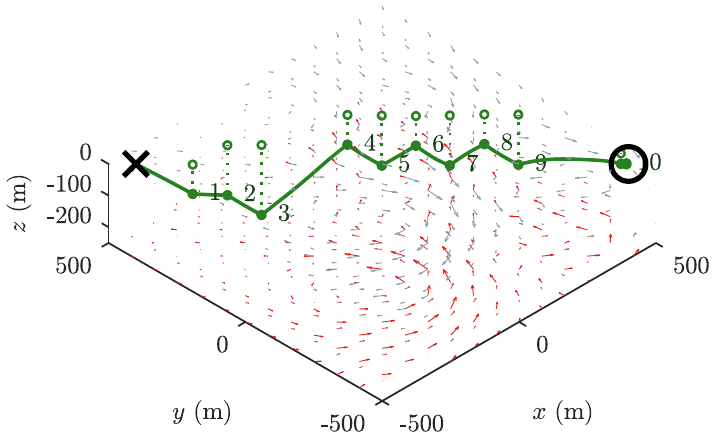}
            \label{fig:complex_sparse_iso}
        }\\
        \vspace{-2ex}
        \subfloat[Top view]{
            \includegraphics[width=0.9\linewidth]{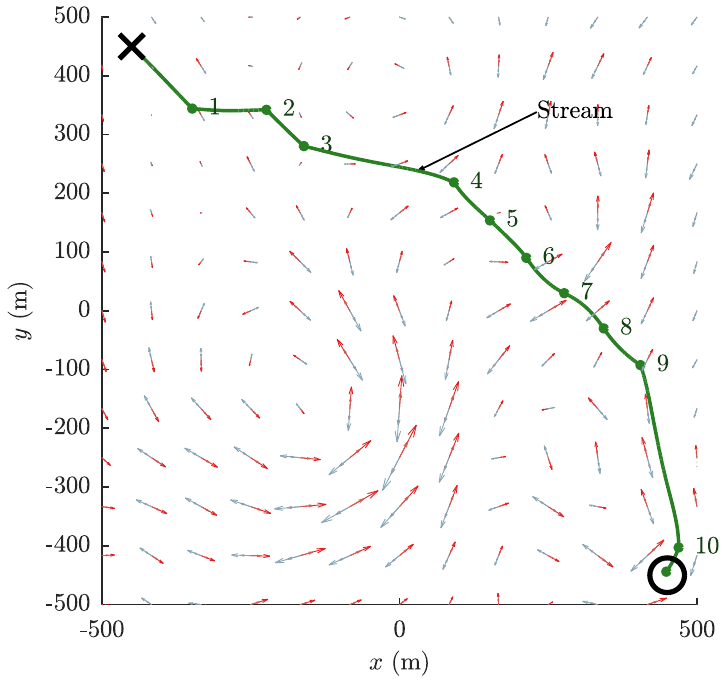}
            \label{fig:complex_sparse_xy}
        }\\
        \vspace{-2ex}
        \subfloat[Depth profile over execution time]{
            \includegraphics[width=0.9\linewidth]{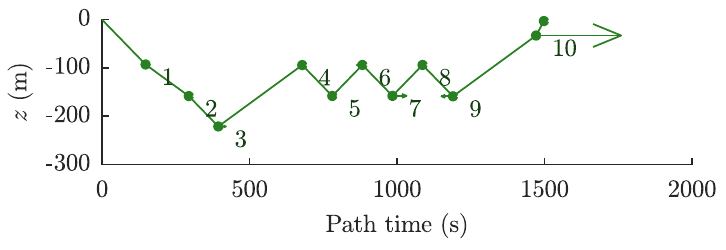}
            \label{fig:complex_sparse_depth}
        }
    \end{minipage}%
    \begin{minipage}{0.5\textwidth}
        \centering
        \subfloat[Isometric view]{
            \includegraphics[width=0.9\linewidth]{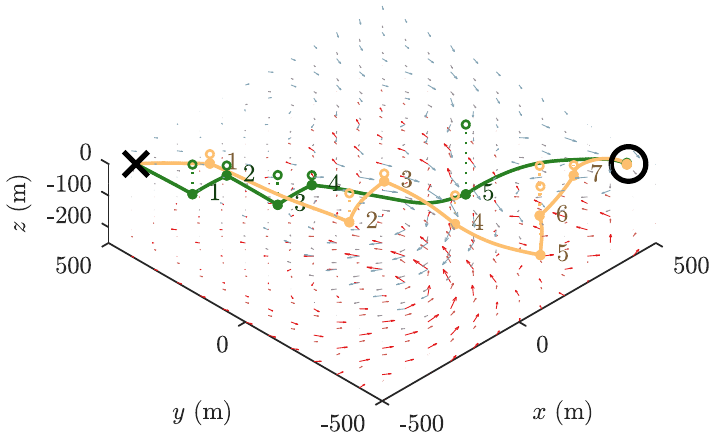}
            \label{fig:complex_denser_iso}
        }\\
        \vspace{-2ex}
        \subfloat[Top view]{
            \includegraphics[width=0.9\linewidth]{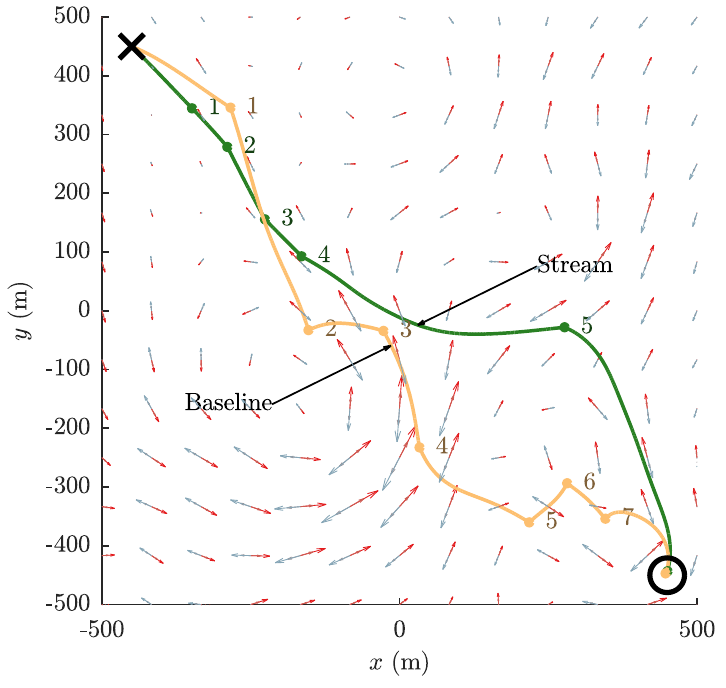}
            \label{fig:complex_denser_xy}
        }\\
        \vspace{-2ex}
        \subfloat[Depth profile over execution time]{
            \includegraphics[width=0.9\linewidth]{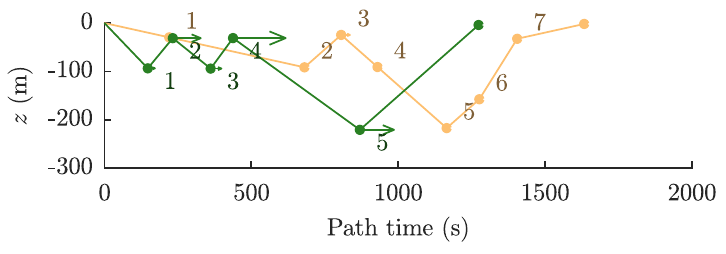}
            \label{fig:complex_denser_depth}
        }
    \end{minipage}
    \caption{
        The proposed and the baseline methods are compared in a simulated 3D flow field, where the glider is to traverse from~$[450, -450, 0]\transpose$ (cross) to~$[450, -450, 0]\transpose$ (circle). The colour of ocean currents goes from light blue to darker red as the depth increases.
        The streamline-based path is shown in green and the baseline path is shown in orange. 
        We used 16 and 54 control samples for (a-c) and (d-f), respectively, and 1024 equispaced 3D position samples for \prm. The arrows in (c) and (f) indicate the magnitude of the current in the direction of glider path.
    }
    \vspace{-3ex}
    \label{fig:complex}
\end{figure*}

\section{Results}
We present two examples to illustrate how the proposed control search algorithm performs against the baseline method.
In the latter, a set of control candidates is sampled from the control surface.
In particular, we compare the methods with different numbers of control candidates and discuss the path quality and the likelihood of finding a solution.
We evaluate each path using the overall time to traverse from the initial position to the goal.

Both methods are implemented in \prm{} to find an optimal sequence of controls from initial position~${\pos_{init} = [-450, 450, 0]\transpose}$ to goal~${\pos_{goal} = [450, -450, 0]\transpose}$.
We sample 1026 position states uniformly in the $xyz$-environment, including the initial position and the goal, to build a roadmap for~\prm{}.
It is important to note that the proposed method can be used in any sampling-based planning algorithms that need to find local edge connections between points. 

Likewise, we have evenly sampled the set of control candidates on the parameterised control line \eqref{eqn:parametrised} and on the control surface (see Remark.~\ref{remark:control_surface}) for our method and the baseline, respectively.
For each control candidate, we forward integrate the control for $125$ steps with time step size~$dt=5$\,\si{s}.
A path is considered to reach the target position when the minimum distance between the path and the target is less than $5$\,\si{m}.
Note that the maximum horizontal speed of the glider is around~$0.9$\,\si{m/s}.

We use a simulated environment in which the spatially-varying ocean current field is incompressible and the $z$-component of velocity is zero (i.e., $w_c = 0$).
In order to show that the proposed algorithm works over challenging environments, the maximum horizontal speed of the flow field is more than double the maximum horizontal speed.



In Fig.~\ref{fig:complex_sparse_iso}-\ref{fig:complex_sparse_depth}, we show the results for both methods with $16$ control samples.
While our method provides a solution (shown in green), the baseline method failed to find a solution.
This is because our method generates far more edge connections ($12423$) than the baseline ($849$) with the same number of control samples.
The number of control candidates was not enough for the baseline to find a path within its roadmap due to a larger search space.
This result 
aligns well with the analysis in Fig.~\ref{fig:analysis}.
The travel time for the path is~$1498$\,\si{s}.

In Fig.~\ref{fig:complex_denser_iso}-\ref{fig:complex_denser_depth}, we have $54$ control samples and both methods were able to find a solution (ours in green, baseline in orange) with $29734$ and~$3738$ edge connections for ours and the baseline, respectively.
The travel time for ours is~$1274$\,\si{s}, whereas that for the baseline is~$1633$\,\si{s}.
Our method performed 12\% better with respect to path quality.

The paths generated using our method exhibit interesting behaviours.
In Fig.~\ref{fig:complex_sparse_iso}-\ref{fig:complex_sparse_depth}, the glider initially dives deep to avoid weak opposing currents and then moves at approximately $150$\,\si{m} depth between $\pos_4$ and~$\pos_9$, where the magnitude of the ocean currents is nearly zero.
The depth profile in Fig.~\ref{fig:complex_sparse_depth} shows the magnitude of the current in the direction of the glider (using arrows), where the glider experiences near-zero currents, except when it nears the goal.
In contrast, the depth profile using our method in Fig.~\ref{fig:complex_denser_depth} shows that the glider is exploiting currents to reach the destination faster which illustrates that our method finds a better path by discovering favourable currents.


As we discussed in Sec.~\ref{sec:analysis}, the proposed method finds a solution with far fewer control samples than the baseline.
This is an important property in practice, since future path computation can be performed during the execution of the current path in a plan-as-you-go manner.
For example, using $k_{PRM}=27$ nearest neighbours suggested in~\cite{star_Karaman2010}, the computation time was $399$\,\si{s} while the path travel time was $1530$\,\si{s}.
This property allows the glider to re-plan when it receives a new mission, or if it needs to adapt to environment changes, without affecting the operation.
In contrast, the baseline method was not able to find any solution with the same number of neighbours.


It is important to note that the edges between each position samples are not straight, since each edge is generated by forward integrating a control under the influence of ocean currents. The curved edge connection is clearly visible between~$\pos_4$ and~$\pos_6$ using our method in Fig.~\ref{fig:complex_denser_iso}-\ref{fig:complex_denser_depth}.

\section{Conclusion and future work}

In this paper, we considered the edge connection problem for underwater gliders in the presence of 2.5D oceanic currents, where we account for non-linear glider dynamics in 3D position space.
We addressed the inherent computational bottleneck by reducing the control search space from 2D to 1D using streamlines.
We showed that the path quality improved significantly for the same number of control samples compared to a baseline method.
We presented two simulated examples that illustrate the improved path quality. We also argued that the proposed method is efficient enough for online re-planning.

Although we showed that the depth-averaged assumption is valid for ocean currents in which gliders operate, this may not be true for other oceanic applications.
One interesting idea is to extend the method to consider a full 3D flow field in order to consider applications such as underwater vehicles operating in shallow water, or other cases where the vertical component may be non-negligible.
We would also like to consider solving for discrete forecast data points using GP regression~\cite{lee2018active,yoo2016online}, and planning over uncertainty~\cite{Chanyeol2019,yoo2013provably}.
Another interesting practical application is to exploit the re-planning capability with an online current mapping algorithm~\cite{brian2019online} where ocean estimation changes over time.


\balance


\bibliographystyle{IEEEtran}
\bibliography{ref}

\end{document}